%% file: main_2026_v5_sign_corrected.tex
\documentclass[9pt]{article}
\usepackage{spconf,amsmath,graphicx}

\usepackage[utf8]{inputenc} \usepackage{enumitem}
\usepackage{IEEEtrantools}
\usepackage{url}
\usepackage[normalem]{ulem}
\usepackage[T1]{fontenc}
\usepackage{url}
\usepackage{ifthen}
\usepackage{multirow}
\usepackage{url}
\usepackage{graphicx,color,epsfig,rotating,subfigure}       
\usepackage{algorithm,algcompatible,xcolor,cite}
\usepackage{graphicx,color,epsfig,rotating,subfigure}
\usepackage{amsfonts,amsmath,amssymb, bm}
\usepackage{algorithm}
\usepackage{subfigure}
\usepackage{algpseudocode}

 \newcommand{\um}[1]{}    
 \newcommand{\jm}[1]{} 
 \newcommand{\cn}[1]{}
 \newcommand{\km}[1]{}

\newcommand{\fix}[1]{#1} 
\input{macros}

\usepackage{graphicx}
\usepackage{tabularx,booktabs,hhline}
\usepackage{hyperref}
\usepackage{color, colortbl}
\definecolor{LightCyan}{rgb}{0.88,1,1}
\definecolor{lightgray}{gray}{0.95}
\usepackage{multirow}
\usepackage{soul} 
\usepackage{amsthm}

\usepackage{url}

\usepackage{breakurl}
\def \mM \mathbf{M}
\def \mQ \mathbf{Q}
\def \mS \mathbf{S}
\def \mV \mathbf{V}
\def \mU \mathbf{U}
\def \mX \mathbf{X}

\newtheorem{definition}{Definition}
\newtheorem{lemma}{Lemma}
\newtheorem{corollary}{Corollary}

\newtheorem{proposition}{Proposition}

\renewcommand{\subsubsection}[1]{\noindent {\bf #1. }}

\newcommand{\Real}{\mathbb{R}}
\newcommand{\norm}[1]{\left\|{#1}\right\|}
\newcommand{\Exp}[2]{E_{#1}\left[{#2}\right]}
\newcommand{\Prob}[1]{P\left[{#1}\right]}
\newcommand{\Probd}[2]{P_{#1}\left[{#2}\right]}
\newcommand{\sD}{\mathcal{D}}
\newcommand{\sC}{\mathcal{C}}
\newcommand{\indic}[1]{1\left\{{#1}\right\} }

\newcommand{\pdd}[2]{\tilde{F}_{#1}\left( {#2}\right)}
\usepackage{stackrel}

\makeatletter
\def\blfootnote{\gdef\@thefnmark{}\@footnotetext}
\makeatother

\title{Pairwise distortion distribution  for compression and quantization}

\name{\vspace{-0.1in} Carl Nuzman$^{\star}$, Jeongmin Chae$^{\dagger}$, Kursat Rasim Mestav$^{\star}$,
 and Urbashi Mitra$^{\dagger}$}
 \vspace{0.5in}
 
\address{$^\star$ Nokia Bell Labs\\$^\dagger$ Department of Electrical  \& Computer Engineering, University of Southern California
}

\begin{document}
%
\maketitle

\vspace{-0.1in}
\begin{abstract}
\vspace{-0.1in}
Choosing the latent dimension is a critical challenge in data compression, where conventional hyperparameter optimization is costly and ad hoc. Intrinsic dimension provides a good alternative, which connects information-theoretic notions such as Renyi information dimension (RID) to rate-distortion dimension and random codebook compression. In this work, we introduce the \emph{pairwise distortion distribution (PDD)} and its asymptotic limit, \emph{pairwise dimension (PD)}, which generalize the correlation integral to general distortion measures. We show that PD bounds RID, and PDD provides a lower bound for random codebook quantization performance, which can guide a practical use for compression, quantization, and dimensionality reduction applications.
\end{abstract}
\vspace{-0.05in}

\blfootnote{Funded by one or more of the following grants: ARO W911NF1910269, ARO W911NF2410094,  NSF CIF-2311653, NSF CIF-2148313, NSF RINGS 2148313,  ONR N00014-22-1-2363, NSF DBI-2412522, and Resilient \& Intelligent NextG Systems (RINGS) program}

\vspace{-0.15in}

\section{Introduction}\label{sec:intro} 
\vspace{-0.1in}
Many applications such as representation learning, compression, and quantization rely on data having an intrinsic dimension much lower than the raw, or extrensic dimension. 
\cite{andonie2019hyperparameter,liao2022empirical,campadelli2015intrinsic,berchtold2000independent}.Machine learning is a useful tool for finding such representations, in which the assumed intrinsic dimension is a key hyperparameter \cite{gong2019intrinsic}.  
Conventional approaches address this by exhaustive search, which are computationally expensive and provides little insight on the geometry of the data \cite{bergstra2012random,hutter2019automated}.

To address this, a large body of work focuses on estimating the intrinsic dimension (ID) of sets and of probability distributions, under various definitions \cite{KawaDembo1994}.   Estimators for sets include maximum likelihood estimation (MLE) \cite{levina2004maximum}, TWO-NN \cite{facco2017estimating}, and PCA-based approaches \cite{camastra2016intrinsic}, which leverage the local or global data statistics.  
Two well known definitions for ID are the Hausdorff dimension and the box counting dimension \cite{camastra2016intrinsic,billingsley1960hausdorff,li2009improved}.
The information dimension, or the Hausdorff information dimension (HID), is the smallest Hausdorff dimension among sets having probability 1.

In information theory, an important quantity is the Renyi information dimension (RID) \cite{renyi1961measures,wu2010renyi}. The RID of a random vector $X\in\Real^n$, for $m>0$, is defined in terms of a series of quantized versions $\left\langle X\right\rangle_m = \frac{\left\lfloor mX \right\rfloor }{m}$ as
$d_I(X) = \lim_{m\to\infty} \frac{H\left( \left\langle X\right\rangle_m\right)}{\log m}$
when the limit exists, where $H$ is the Shannon entropy.
The RID is connected in \cite{KawaDembo1994} to the asymptotic scaling of the rate distortion curve as distortion approaches zero, for a wide class of distortion measures. 
In \cite{GrassProc1983}, the authors introduce the correlation integral and corresponding correlation dimension as computationally simpler alternatives to HID for quantifying the dimension of a distribution.

In this paper, we make connections between the correlation integral, the RID, and compression by generalizing the correlation integral from Euclidean distances to a general distortion measure. To this end, we introduce a new function called \emph{the pairwise distortion distribution (PDD)}, whose asymptotic limit defines a distortion-dependent notion of dimension referred to \emph{pairwise dimension (PD)}. Both PDD and PD are relatively easy to estimate from data and to derive analytically for known distributions.

We show that when distortion is a power of a vector norm, the PD gives a \fix{lower bound} on the RID, and thus on the asymptotic behavior of the rate-distortion curve for a wide class of distortion functions. 
A simple PDD-based computation \fix{also} yields a lower bound on the quantization performance of random quantization codebooks. Taken together, these properties suggest that the PDD-derived rate-distortion curve can be a good guide to the quantization performance, which we confirm via empirical experiments.


\vspace{-0.2in}
\section{Background}\label{sec:revprelim}
\vspace{-0.1in}
We review the correlation integral and correlation dimension, which are the key concepts for our framework. The definition below is modified from, but equivalent to, the version given in \cite{GrassProc1983}.

\begin{definition}[Correlation Integral and Correlation Dimension {\cite{GrassProc1983}}]
Let $X \sim \mathcal{D}$ be a random vector on $\mathbb{R}^N$.  
For $r \geq 0$, the \emph{correlation integral} is defined as
\begin{equation}
\vspace{-0.02in}
    C(r) \;=\; P_{X_1, X_2 \,\sim\, \mathcal{D}} \!\left[ \, \|X_1 - X_2\| \leq r \,\right],\vspace{-0.02in}
\end{equation}
where $X_1$ and $X_2$ are independent realizations of $X$. \km{Should we mention this definition is slightly modified but equivalent to the definition in the citation [8]?}\jm{Maybe we cite [8] as well?} \km{Oh, I actually meant [11], Grassberger Procaccia definition} \cn{Done }
The \emph{correlation dimension} of $\mathcal{D}$, if the limit exists, is given as
\begin{equation}
    d_C(X) \;=\; \lim_{r \to 0} \frac{\log C(r)}{\log r}.
\end{equation} 
\end{definition}
$C(r)$ is the cumulative distribution function of the random variable $Z \;=\; \|X_1 - X_2\|$,
which represents the distance between two independent samples drawn from $\mathcal{D}$. In particular, if $X$ has an absolutely continuous distribution supported on a $d$-dimensional subset of $\mathbb{R}^N$, then $d_C(X) = d$.

{While optimal codebooks achieve rate-distortion limits, its structure is difficult to characterize. Instead, we focus on the random codebook quantization scheme (RCQ), which provides a mathematically tractable baseline for PDD analysis. }
{
Let $\mathcal{C} = \{c_1, \dots, c_L\} \subset \mathbb{R}^N$ denote a quantization codebook with $L$ codewords. For a distortion measure $\rho(x,y)$,  a quantizer maps an input $X \sim \mathcal{D}$ to its nearest codeword, $j(X) = \arg\min_{n} \rho(X, c_n)$. The mean distortion of the codebook is $\mu(\mathcal{C}) = \mathbb{E}_{X \sim \mathcal{D}}\Big[\min_{n} \rho(X, c_n)\Big].$
When the codebook $\mathcal{C}$ is generated randomly, with each codeword drawn independently from $\mathcal{D}$, the expected distortion over random codebooks is $\mu_L = \mathbb{E}_{\mathcal{C}}\big[\mu(\mathcal{C})\big].$}




\vspace{-0.15in}
\section{PDD-based framework}
\vspace{-0.1in}
We introduce a new distribution and its corresponding notion of dimension, which generalize the correlation integral by replacing the Euclidean distance with an arbitrary distortion measure $\rho(x,y)$. 

\begin{definition}[Pairwise Distortion Distribution and Pairwise Dimension]
Let $X \sim \mathcal{D}$ be a random vector on $\mathbb{R}^N$, and let 
$\rho:\mathbb{R}^N \times \mathbb{R}^N \to \mathbb{R}_{\ge 0}$ 
be a distortion measure. {For $r \geq 0$,}
the \emph{pairwise distortion distribution} (PDD) of $\mathcal{D}$ with respect to $\rho$ is defined as
$\pdd{\mathcal{D},\rho}{r} 
\;=\; P_{X_1,X_2 \sim \mathcal{D}}\!\left[ \rho(X_1,X_2) \leq r \right].$
The corresponding notion of dimension, the \emph{pairwise dimension} (PD), is defined as
\begin{equation}
    \vspace{-0.03in}
    d_P(\mathcal{D},\rho) 
    \;=\; \lim_{r \to 0} 
        \frac{\log \pdd{\mathcal{D},\rho}{r}}{\log r}.
\end{equation}

\end{definition}
We now develop a tool for compression analysis, built upon two key fundamental relationships: (i) between the pairwise distortion distribution (PDD) and the Renyi information dimension (RID), and (ii) between PDD and random codebook quantization (RCQ).
\vspace{-0.1in}
\subsection{Connection between PDD and RID}

Proposition 4 in \cite{WuVerdu2010}
 shows  an alternative representation for the RID that is closely related to PDD. For $p$-norm on $\Real^n$ with $1\leq p \leq \infty$,
 $$\vspace{-0.02in}
 d_I(X) = \lim_{r\to 0} \frac{\Exp{X_1}{\log \Probd{X_2}{\norm{X_1-X_2}_p\leq r}}}{\log r}
 $$
This connects the RID to probabilities of balls of size $r$ around random centers. \km{We could also mention how correlation distance or PDD is related to balls earlier to connect it with this. For example, the correlation integral is searching for neighbors in a ball.} \cn{Maybe leave out, due to space.} We show that PDD provides \fix{a lower bound} for RID.
\begin{proposition}\label{prop:dim_bound}
    For distortion functions of the form $\rho(x,y)=\norm{x-y}_p^\nu$, for $1 \leq p \leq \infty$ and $\nu > 0$, if both limits exist,
 $$
 \nu \, d_P(\mathcal{D},\rho) \,\fix{\leq}\, d_I(X).
 $$
\end{proposition}
{\noindent\em Proof}: 
For each $r$,
\begin{IEEEeqnarray*}{rCl}
\IEEEeqnarraymulticol{3}{l}{E_{{X_1}}{\log \Probd{X_2}{\norm{X_1-X_2}_p \leq r}}} \\ \quad
 & = & 
\Exp{X_1}{\log \Exp{X_2}{1\left\{ \norm{X_1-X_2}_p\leq r\right\}}} \\
& \leq  &
\log \Exp{X_1}{ \Exp{X_2}{1\left\{ \norm{X_1-X_2}_p\leq r \right\}}} \\
& = &
\log \Probd{X_1, X_2}{\rho(X_1,X_2)\leq r^\nu}
\; = \; \log \pdd{\mathcal{D},\rho}{r^\nu}
\end{IEEEeqnarray*} 

\fix{Taking $r<1$ so that $\log r < 0$, dividing through by  $\nu \log r =  \log r^\nu$ on both sides, and taking the limit as $r\to 0$ gives the result. \qed}

The \emph{ rate-distortion dimension} of a distribution is defined in \cite{KawaDembo1994} for distortions of the form $\rho(x,y) = \norm{x-y}^\nu$, as
$$
\vspace{-0.02in}
d_R(\mathcal{D}) = \lim_{\epsilon \to 0} \frac{R_\mathcal{D}(\epsilon^\nu)}{-\log(\epsilon)}\vspace{-0.02in}
$$ 

where $R_\mathcal{D}(D)$ is the Shannon rate-distortion function for distribution $\mathcal{D}$.
\km{Is $R_\mathcal{D}$ defined in the citation? We don't define it here.} \cn{fixed}
The authors show that $d_R(\mathcal{D}) = d_I(\mathcal{D})$ for any $\nu > 0$ and any norm equivalent to $\norm{}_\infty$ (including $\norm{}_p$ for any $1\leq p \leq \infty$). Combining this result with Proposition~\ref{prop:dim_bound}, 
we see that the asymptotic increase of $R_{\mathcal{D},\rho}(D)$ for small distortion $D$ can be \fix{lower} bounded based on $d_P(\mathcal{D},\rho)$.
Thus, the PDD curve near $r=0$ yields useful information about the effective dimension of a distribution from a rate-distortion point of view.

\vspace{-0.15in}
\subsection{Connection between PDD and RCQ}
\vspace{-0.1in}
We next show a connection between the PDD and the performance of random codebook quantization. 


\begin{proposition}\label{prop:pdd_rcq}
Let $X \sim \mathcal{D}$ on $\mathbb{R}^N$ be a random vector. Generate a quantization codebook by drawing $L$ codewords $\left\{C_1,\ldots, C_L\right\}$ independently from $\mathcal{D}$. For a given distortion function $\rho(x,y)$, the expected distortion over random codebooks, $\mu_L = \mathbb{E}_{C \sim\mathcal{D}^L}\Big[\mathbb{E}_{X \sim \mathcal{D}}\Big[\min_{n} \rho(X, C_n)\Big]\Big],$ satisfies 
\begin{align}\label{eq:lower_bound}
    \mu_L \geq \int_0^\infty \left( 1-\pdd{\mathcal{D},\rho}{r}\right)^{L}\; dr.
\end{align}
\end{proposition}
{\noindent\bf Remark.} $\pdd{\mathcal{D},\rho}{r}$ can be computed or bounded analytically for simple distributions (See e.g. Corollary~\ref{cor:gaussian} and \ref{cor:sparse}) or estimated empirically from sufficiently many samples from a dataset. 
\cn{ Let's try to move the following sentence to somewhere else, where we should say more about how the bound can be useful} 

\jm{I think we can write proof sketch instead in case we need space.} \cn{I agree}

\noindent \emph{Proof:} 
For given $x\in\Real^N$, define non-negative random variables $Y_n(x)=\rho({x},{C_n})$ with $C_n$ drawn independently from $\mathcal{D}$.
Denoting the CDF of $Y_n(x)$ as $F_{Y}(r;x)$, the  CDF of $\min_n Y_n(X)$ is $1-\left(1-F_{Y}(r;x)\right)^L$. We have
\begin{align}
 \Exp{\sC}{\min_n Y_n(r;x)}\nonumber &=  \int_0^{\infty} \left( 1-F_{Y}(r;x)\right)^L \, dr \nonumber\\
& =  \int_0^\infty \left( \Exp{c\sim\sD}{\indic{\rho({x},{C_1})\geq r}}\right)^L \, dr \nonumber ,
\end{align} thus,
\begin{eqnarray*}
\mu_L 
 & = & \Exp{X\sim\sD}{\int_0^\infty \left(\Exp{C_1\sim\sD}{\indic{\rho(X,C_1)\geq r}}\right)^L \, dr } \\
&\geq&\int_0^\infty \left( 1-P\left[\rho(X,C_1)\leq r\right]\right)^{L}\; dr,
\end{eqnarray*}
where the second line uses Jensen's inequality.\qed

\vspace{-0.2in}
\subsection{Additional analysis}
\vspace{-0.1in}
The pairwise dimension of $d_P(\mathcal{D},\rho)$ equals $d$ when 
$
\pdd{\mathcal{D},\rho}{r} \approx \beta r^d
$
for small $r$. More generally, bounds on the behavior of the PDD near $r=0$ bound the 
behavior of the distortion $\mu_L$ of the random codebook as $L\to\infty$. 
\vspace{-0.1in}
\begin{lemma}\label{lemma:alpha}
Let $X \sim \mathcal{D}$ on $\mathbb{R}^{N}$and suppose that $\pdd{\mathcal{D},\rho}{r} \leq \beta r^{\alpha}$ for some $\alpha>0$, $\beta > 0$. Then
$$
\mu_L \geq \alpha^{-1} \beta^{-1/\alpha} B(L+1,\alpha^{-1}),
$$
\km{where B is Beta function.}
\end{lemma}
In particular, $\mu_L$ scales as $(L+1)^{-1/\alpha}$ for large $L$.

\noindent\emph{Proof.}
\begin{align}
    \mu_L &\geq \int_0^\infty \left( 1-\pdd{\mathcal{D},\rho}{r}\right)^{L}\; dr\\
    &\geq \int_0^{\beta^{-1/\alpha}} \left( 1-\beta r^{\alpha}\right)^{L}\; dr \\
    &= \alpha^{-1} \beta^{-1/\alpha}\int_{0}^{1}(1-\tau)^{L}\tau^{1/\alpha-1} d\tau\label{eq:lb_beta}
\end{align} 
\km{[Edited] We could use equal sign in steps 8,9.}\jm{Agree. 8, 9 can be equality?}
When the distortion function is a function of the difference between its arguments, $\rho(x,y) = f(x-y)$, and when $X$ has an absolutely continuous and symmetric distribution on $\Real^N$ with pdf $p_X(x)=p_X(-x)$, the bound on random codebook quantization can further be related to the volumes of balls in $\Real^N$ under the given distortion function.

\begin{proposition}\label{prop:sym} Let $X\in\Real^N$ have absolutely continuous distribution $\mathcal{D}$ with density $p_X(x)=p_X(-x)$. Define distortion function $\rho(x,y)=f(x-y)$ for non-negative integrable $f$. Then if $\Exp{\mathcal{D}}{p_X(X)} < \infty$, 
the density $p_Y$ of $Y = X_1-X_2$ satisfies $p_Y(0) = \Exp{\mathcal{D}}{p_X(X)}$, and we have
$$
\pdd{\mathcal{D},\rho}{r} \leq p_Y(0)\, V_f(r) 
$$
where $V_f(r)$ is the volume of the set $f(x)\leq r$.
\end{proposition}
\noindent\emph{Proof.}
Defining $Y = X_1-X_2$ for iid $X$, the distribution of $Y$ is of the form $p_Y(y) = \int_{\Real^N} p_X(x)p_X(y-x) \, dx = \int_{\Real^N} p_X(x)p_X(x-y)\, dx$. 
Considering the integral of non-negative quantity $(p_X(x)-p_X(x-y))^2$ over $\Real^N$, establishes that $p_Y(y) \leq p_Y(0) = \Exp{X}{p_X(X)}$ if that expectation exists.
Finally, $\pdd{\mathcal{D},\rho}{r} = \Prob{f(Y)\leq r} \leq p_Y(0) V_f(r)$. \qed

Combining Proposition~\ref{prop:sym} and Lemma~\ref{lemma:alpha}, the pairwise dimension and asymptotic quantization behavior can be characterized in terms of the scaling of volumes in $\Real^N$ under different distortion measures.

For example, for a distortion measure $\rho(x,y) = \norm{x-y}_p^\nu$, we have $V_f(r) = \mbox{Vol}_p(r^{1/\nu})$ where $\mbox{Vol}_p(r)$ is the volume of the $p$-norm ball in $\Real^N$.
For mean-squared error distortion $\rho(x,y)=\Vert x-y \Vert_2^2$, in particular, we have
\begin{align}
\pdd{\mathcal{D},\rho}{r} \leq \;& p_Y(0)\, \mbox{Vol}_2(r^{1/2})
  =  p_Y(0) \cdot \frac{\pi^{N/2}}{\Gamma(N/2+1)}r^{N/2},\nonumber
\end{align}
\km{where $\Gamma$ is the Gamma function.}
In the following, we analytically compute the corresponding lower bound for various types of distributions for $\mathcal{D}$ on $\mathbb{R}^{N}$.
\begin{corollary}[Multi-variate Gaussian distribution]\label{cor:gaussian} Suppose  $\mathcal{D} \sim \mathcal{N}(0,\Sigma)$. Then, 
\begin{align}
    \mu_L \geq \frac{8}{N}\left(\Gamma(N/2+1)\right)^{2/N} \mbox{det}(\Sigma)^{1/N} B(L+1,2/N).\nonumber
\end{align}

\end{corollary}

\emph{Proof:}
The difference of two independent
variables from the distribution is distributed as $\mathcal{N}(0,2\Sigma)$. Then, $p_Y(0)$ is 
$(4\pi)^{-N/2}\; \mbox{det}(\Sigma)^{-1/2}$, and applying Lemma~\ref{lemma:alpha} gives the result. \qed

This MSE lower bound scales with $\mbox{det}(\Sigma)^{1/N}$, i.e., the geometric mean of the eigenvalues of $\Sigma$.

\begin{figure*}[t!]
\centering
\includegraphics[width=16cm]{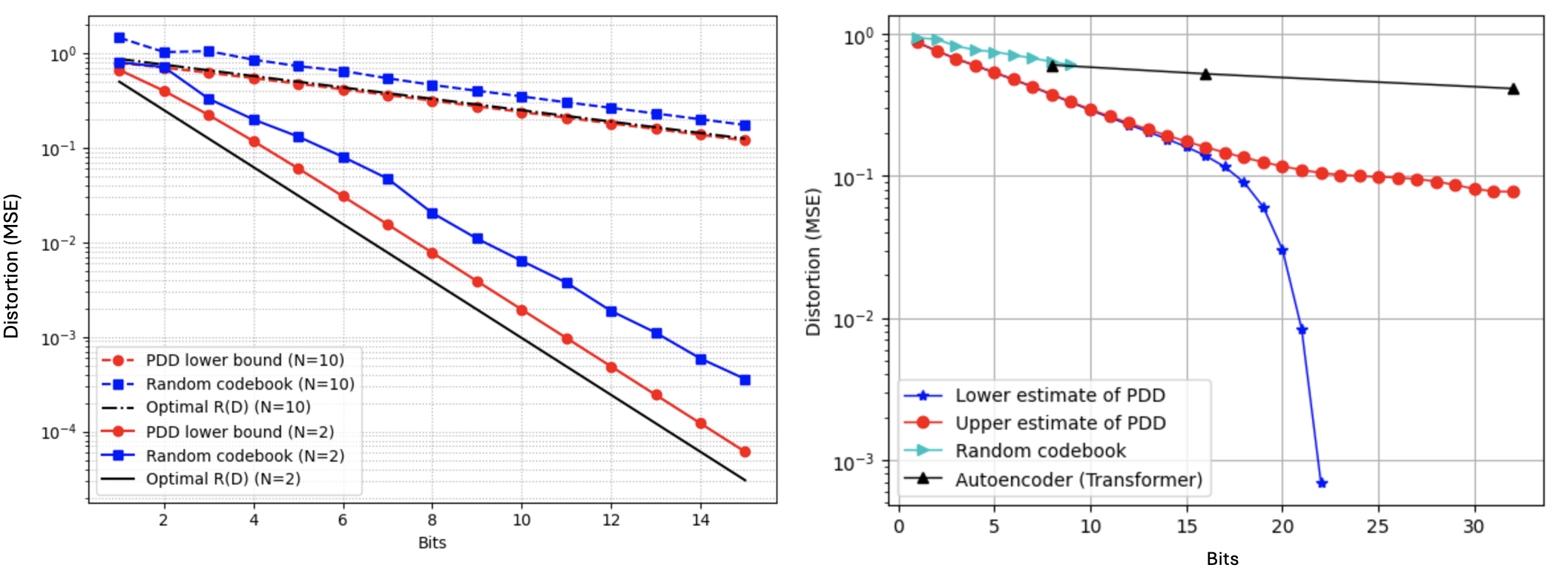}
   \vspace*{-0.2in}
    {\caption{Rate-distortion performance comparision for Gaussian normal vectors and MNIST data.}}
\vspace*{-0.1in}
\end{figure*}


\begin{corollary}[$k$-sparse vectors]\label{cor:sparse} Suppose $X \in \mathbb{R}^{N}$ is a $k$-sparse random vector with $N\geq 2k$, where the $k$ non-zero elements are chosen uniformly at random from the set $\mathcal{S}=\{1,\dots,N\}$, and the non-zero coefficients are independent and identically distributed as $\mathcal{N}(0,\sigma^2)$. Then under MSE distortion,
$\mu_L \geq \frac{8\sigma^2}{k}\binom{N}{k}\Gamma\left(\frac{k}{2}+1\right)^{2/k} B(L+1,2/k).$
\end{corollary}

\emph{Sketch proof:}
The supports of two variables $X_1$ can intersect on anywhere from $0$ to $2k$ elements. Enumerating these possibilities, one can show that 
{
\begin{align}
    \pdd{\mathcal{D},\rho}{r} &\geq\frac{1}{\binom{N}{k}}\cdot\frac{\gamma(k/2,r/4\sigma^2)}{\Gamma(k/2)} + O\left(r^{(k+1)/2}\right)\nonumber
\end{align}}
Considering the behavior of $\gamma(m,r)$ for small $r$ and applying Lemma~\ref{lemma:alpha} yields the result. \qed

\vspace{-0.15in}
\subsection{Empirical estimation of the PDD}
\vspace{-0.05in}
Part of the motivation for using the PDD  is that it is relatively easy to estimate from data.
When data is limited to a relatively small number  $K$ of samples $x_i$, it can make sense to use all $K(K-1)$ pairs (as in \cite{GrassProc1983}), $\hat{F}(r) = \frac{1}{K(K-1)} \sum_{i=1}^K \sum_{j\neq i} 1\left\{\rho(x_i,x_j) \leq r \right\}.
$
When many data samples are cheaply available, $2K$ samples can be used to form the estimate
\begin{equation}\label{eqn:emp_est}
\vspace{-0.02in}
\hat{F}(r) = \frac{1}{K} \sum_{i=1}^K 1\left\{\rho(x_{2i-1},x_{2i}) \leq r \right\}
\vspace{-0.02in}
\end{equation}
which converges faster for a given number of distortion evaluations and is more easily analyzed, as the sum of $K$ independent Bernoulli random variables. \km{These $x_{2i-1}, x_{2i}$ are different than above equation, right? There are K samples, but we get up to 2K here by random pair selection?} \cn{Yes. Modified text to clarify.}
For this estimator, the coefficient of variation is $\sqrt{\frac{1-\pdd{\rho}{r}} {\pdd{\rho}{r} K}}$, from which confidence intervals may easily be computed. To estimate the PDD accurately for small values of $r$, we need $K\geq 1/\pdd{\rho}{r}$ pairs of samples. Future work will consider using methods such as importance sampling to increase the efficiency of estimating small values of the PDD.

\vspace{-0.15in}
\section{Numerical results}\label{sec:numerical}
\vspace{-0.3cm}
We illustrate the proposed framework on both synthetic and empirical datasets to begin to address two key questions:  
\emph{(1) Does PDD provide a meaningful lower bound on quantization error?} and \emph{(2) Can PDD be used to estimate an appropriate latent dimension for data compression?}
For the synthetic case, {independent} gaussian vectors drawn from $\mathcal{N}(0,1)$ are generated with an ambient dimension of $N=2, 10$. For the empirical dataset, we employ the MNIST database of handwritten digits~\cite{lecun2010mnist}, representing each data sample with dimension $N=784$. 

\subsubsection{Synthetic Data: Gaussian Vectors}
We compare against two baselines: the random codebook scheme described in Section~\ref{sec:revprelim} and the optimal rate–distortion curve~\cite{max1960quantizing,cover2006elements}.  
For the random codebook scheme, we generate a codebook $\mathcal{C}$ of size $L \times N$, where $L = 2^b$ and $b$ denotes the bit rate. Each codeword is constructed as a gaussian vector drawn from $\mathcal{N}(0,1)$. For each bit rate $b \in [1,15]$ and $L=2^{b}$, we analytically compute PDD in Corollary~\ref{cor:gaussian} with parameters $N=2, 10$, and $\Sigma=I$. The optimal rate–distortion curve for a Gaussian distribution is the classical result from Shannon’s rate–distortion theory\cite{cover2006elements}, that is $D(R)=\sigma^2 2^{-2b/N}$.

\subsubsection{Empirical Data: MNIST with ViT-Based Autoencoder}
We implement a Vision Transformer–based autoencoder (ViT-AE)~\cite{vaswani2017attention,han2021transformer} tailored for MNIST images of size $28\times 28$. Each image is divided into 28 patches of length 28, which are projected into a 128-dimensional embedding space. The encoder consists of 5 Transformer layers with 16 attention heads and MLP blocks of hidden size 256, producing a compact latent representation. For the latent dimension $d$, we experiment with multiple settings, specifically $d \in \{8, 16, 32\}$. This latent vector is quantized using a 2-bit uniform quantizer, restricting activations to four discrete levels, and then decoded through a symmetric Transformer to reconstruct the input while achieving efficient compression. Next, we compute the lower and upper bounds of the PDD framework.
\cn{Is this revised description ok?}\jm{Yes.}
We sampled $K = 10^6$ pairs of points from the MNIST dataset and estimated the PDD for various values of $r$ using \eqref{eqn:emp_est}. The 95\% upper and lower confidence intervals of the PDD estimate were then integrated to obtain upper and lower estimates of the right side of \eqref{eq:lower_bound}. \cn{Should we change the legend in the figure to "Upper estimate of PDD bound" and "Lower estimate of PDD bound"?} \jm{Yes, fixed.}



\subsubsection{Performance analysis}
Fig.~1 illustrates rate-distortion (MSE) curves for gaussian vectors and MNIST dataset. 
In the case of gaussian vectors, on a $\log D(R)$ vs bits scale, the optimal R(D) curves decay linearly with slope $-2/N$. The slope directly reflects the intrinsic dimension of the gaussian data, which is $N$. Thus, when $N=2$, the curve shows steeper slope than $N=10$. The PDD curves are almost parallel to the optimal R(D) curves, consistent 
with Proposition~\ref{prop:dim_bound}.
This confirms that PDD captures the correct asymptotic scaling behavior. In addition, PDD provides a lower bound for the random codebook scheme, in alignment with Proposition~\ref{prop:pdd_rcq}. 

Next, we introduce our method for identifying an appropriate latent dimension using PDD by leveraging its MSE performance estimate. For the MNIST dataset, the upper bound (ub) of PDD curve provides an accurate approximation of the MSE observed with a latent dimension $d \in \{8,16,32\}$ in the ViT-based autoencoder, demonstrating its effectiveness as a performance predictor. 
{{One practical use of MSE estimate is predicting achievable distortion at a given bit rate. For example, with 20 bits (corresponding to a latent dimension at least 10), we expect MSE at least 0.1 for compression models. The gap between ViT-AE performance and PDD estimates suggests that the predictive MSE is more accurate at lower bits (smaller latent dimension), while at higher bits, we see room for improvements in ViT-AE. This predictive capability is powerful, as it enables us to anticipate both the latent dimension and the corresponding MSE prior to training. }}



\vspace{-0.2in}
\section{Conclusions}
\vspace{-0.1in}
We introduced the pairwise distortion distribution (PDD) as a novel proxy to analyze data geometry, intrinsic dimension, and compression performance. We demonstrated that (1) PD provides \fix{a lower bound} for Renyi information dimension (RID) and (2) PDD yields a tractable lower bound on the random codebook quantization error, providing the direct performance estimate without model training. Our numerical results on gaussian vectors and MNIST dataset show that PDD can capture the right intrinsic scaling behavior as RID and predicts achievable distortion, providing a practical framework to estimate a good latent dimension for autoencoders.

\newpage

\bibliographystyle{IEEEbib}
\bibliography{ref_id}
\end{document}

%% file: macros.tex
\long\def\comment#1{}

\newfont{\bbb}{msbm10 scaled 700}

\newfont{\bb}{msbm10 scaled 1100}

\renewcommand{\arg}{{\hbox{arg}}}


%% file: ref_id.bib
@article{facco2017estimating,
  title={Estimating the intrinsic dimension of datasets by a minimal neighborhood information},
  author={Facco, Elena and d’Errico, Maria and Rodriguez, Alex and Laio, Alessandro},
  journal={Scientific reports},
  volume={7},
  number={1},
  pages={12140},
  year={2017},
  publisher={Nature Publishing Group UK London}
}

@misc{lecun2010mnist,
  title        = {The {MNIST} Database of handwritten digits},
  author       = {LeCun, Yann and Cortes, Corinna and Burges, Christopher J.C.},
  howpublished = {\url{http://yann.lecun.com/exdb/mnist/}},
  year         = {2010}
}

@article{max1960quantizing,
  title={Quantizing for minimum distortion},
  author={Max, J.},
  journal={IRE Transactions on Information Theory},
  volume={6},
  number={1},
  pages={7--12},
  year={1960},
  publisher={IEEE}
}

@article{vaswani2017attention,
  title={Attention is all you need},
  author={Vaswani, Ashish and Shazeer, Noam and Parmar, Niki and Uszkoreit, Jakob and Jones, Llion and Gomez, Aidan N and Kaiser, {\L}ukasz and Polosukhin, Illia},
  journal={Advances in neural information processing systems},
  volume={30},
  year={2017}
}

@article{han2021transformer,
  title={Transformer in transformer},
  author={Han, Kai and Xiao, An and Wu, Enhua and Guo, Jianyuan and Xu, Chunjing and Wang, Yunhe},
  journal={Advances in neural information processing systems},
  volume={34},
  pages={15908--15919},
  year={2021}
}

@article{bergstra2012random,
  title={Random search for hyper-parameter optimization},
  author={Bergstra, James and Bengio, Yoshua},
  journal={The journal of machine learning research},
  volume={13},
  number={1},
  pages={281--305},
  year={2012},
  publisher={JMLR. org}
}

@book{hutter2019automated,
  title={Automated machine learning: methods, systems, challenges},
  author={Hutter, Frank and Kotthoff, Lars and Vanschoren, Joaquin},
  year={2019},
  publisher={Springer Nature}
}

@article{levina2004maximum,
  title={Maximum likelihood estimation of intrinsic dimension},
  author={Levina, Elizaveta and Bickel, Peter},
  journal={Advances in neural information processing systems},
  volume={17},
  year={2004}
}

@article{camastra2016intrinsic,
  title={Intrinsic dimension estimation: Advances and open problems},
  author={Camastra, Francesco and Staiano, Antonino},
  journal={Information Sciences},
  volume={328},
  pages={26--41},
  year={2016},
  publisher={Elsevier}
}

@book{cover2006elements,
  author    = {Thomas M. Cover and Joy A. Thomas},
  title     = {Elements of Information Theory},
  edition   = {2},
  publisher = {Wiley-Interscience},
  year      = {2006}
}

@article{andonie2019hyperparameter,
  title={Hyperparameter optimization in learning systems},
  author={Andonie, R{\u{a}}zvan},
  journal={Journal of Membrane Computing},
  volume={1},
  number={4},
  pages={279--291},
  year={2019},
  publisher={Springer}
}

@article{billingsley1960hausdorff,
  title={Hausdorff dimension in probability theory},
  author={Billingsley, Patrick},
  journal={Illinois Journal of Mathematics},
  volume={4},
  number={2},
  pages={187--209},
  year={1960},
  publisher={Duke University Press}
}

@inproceedings{gong2019intrinsic,
  title={On the intrinsic dimensionality of image representations},
  author={Gong, Sixue and Boddeti, Vishnu Naresh and Jain, Anil K},
  booktitle={Proceedings of the IEEE/CVF Conference on Computer Vision and Pattern Recognition},
  pages={3987--3996},
  year={2019}
}

@inproceedings{berchtold2000independent,
  title={Independent quantization: An index compression technique for high-dimensional data spaces},
  author={Berchtold, Stefan and Bohm, Christian and Jagadish, Hosagrahar V and Kriegel, H-P and Sander, J{\"o}rg},
  booktitle={Proceedings of 16th International Conference on Data Engineering (Cat. No. 00CB37073)},
  pages={577--588},
  year={2000},
  organization={IEEE}
}

@article{li2009improved,
  title={An improved box-counting method for image fractal dimension estimation},
  author={Li, Jian and Du, Qian and Sun, Caixin},
  journal={Pattern recognition},
  volume={42},
  number={11},
  pages={2460--2469},
  year={2009},
  publisher={Elsevier}
}

@article{wu2010renyi,
  title={R{\'e}nyi information dimension: Fundamental limits of almost lossless analog compression},
  author={Wu, Yihong and Verd{\'u}, Sergio},
  journal={IEEE Transactions on Information Theory},
  volume={56},
  number={8},
  pages={3721--3748},
  year={2010},
  publisher={IEEE}
}

@inproceedings{renyi1961measures,
  title={On measures of entropy and information},
  author={R{\'e}nyi, Alfr{\'e}d},
  booktitle={Proceedings of the fourth Berkeley symposium on mathematical statistics and probability, volume 1: contributions to the theory of statistics},
  volume={4},
  pages={547--562},
  year={1961},
  organization={University of California Press}
}

@article{campadelli2015intrinsic,
  title={Intrinsic dimension estimation: Relevant techniques and a benchmark framework},
  author={Campadelli, Paola and Casiraghi, Elena and Ceruti, Claudio and Rozza, Alessandro},
  journal={Mathematical Problems in Engineering},
  volume={2015},
  number={1},
  pages={759567},
  year={2015},
  publisher={Wiley Online Library}
}

@article{liao2022empirical,
  title={An empirical study of the impact of hyperparameter tuning and model optimization on the performance properties of deep neural networks},
  author={Liao, Lizhi and Li, Heng and Shang, Weiyi and Ma, Lei},
  journal={ACM Transactions on Software Engineering and Methodology (TOSEM)},
  volume={31},
  number={3},
  pages={1--40},
  year={2022},
  publisher={ACM New York, NY}
}

@ARTICLE{WuVerdu2010,
  author={Wu, Yihong and Verdú, Sergio},
  journal={IEEE Transactions on Information Theory}, 
  title={Rényi Information Dimension: Fundamental Limits of Almost Lossless Analog Compression}, 
  year={2010},
  volume={56},
  number={8},
  pages={3721-3748},
  doi={10.1109/TIT.2010.2050803}}

@article{GrassProc1983,
  title={Measuring the Strangeness of Strange Attractors},
  author={Grassberger, Peter and Procaccia, Itamar},
  journal={Physica 9D},
  volume={},
  number={},
  pages={189-208},
  year={1983},
  publisher={North Holland}
}

@ARTICLE{KawaDembo1994,
  author={Kawabata, T. and Dembo, A.},
  journal={IEEE Transactions on Information Theory}, 
  title={The rate-distortion dimension of sets and measures}, 
  year={1994},
  volume={40},
  number={5},
  pages={1564-1572},
  doi={10.1109/18.333868}}
